%% file: root.tex
\documentclass[letter, 10 pt, conference]{ieeeconf}  %

\IEEEoverridecommandlockouts                              %

\usepackage{graphics} %
\let\labelindent\relax
\usepackage{skeldoc}
\usepackage{graphicx}
\usepackage{float}
\usepackage{cuted}
\usepackage{capt-of}
\usepackage[nolist]{acronym}
\usepackage{tikz}
\usepackage{stfloats}
\usepackage{subcaption}
\usepackage{tabularx}
\usepackage{multirow}
\usepackage{tabularray}
\usepackage{xcolor}
\usepackage{utfsym}
\usepackage{pgf}
\usepackage{pgfplots}
\usepackage{adjustbox}

\usetikzlibrary{calc}
\usetikzlibrary{shapes}
\usetikzlibrary{arrows.meta}

\usetikzlibrary{patterns}

\definecolor{whitesmoke}{rgb}{0.96, 0.96, 0.96}
\definecolor{tumorange}{rgb}{0.89, 0.45, 0.13}
\definecolor{tumgreen}{rgb}{0.64, 0.68, 0.0}
\definecolor{tumblue}{rgb}{0.0, 0.396, 0.741}

\title{\LARGE \bf
Trajectory Guidance: Enhanced Remote Driving of highly-automated Vehicles
}

\author{Domagoj Majstorovic$^{*, \dagger}$, Simon Hoffmann$^{*}$ and Frank Diermeyer$^{*}$%
\thanks{$^{*}$The authors are with the Institute of Automotive Technology at the Technical University of Munich (TUM), Boltzmannstr. 15, DE-85748 Garching bei M\"{u}nchen, Germany.}%
\thanks{$^{\dagger}$Corresponding author: {\tt\small domagoj.majstorovic@tum.de}}
}
\usepackage{hyperref}
\usepackage{cite}

\begin{document}
\input{content/acronyms.tex}

\newcommand{\krug}[1]{\tikz[baseline=(char.base)]{\node[shape=circle,draw,minimum size=5mm, inner sep=1pt, semithick, fill=white](char){#1}}}
\newcommand{\krugg}[1]{\tikz[baseline=(char.base)]{\node[shape=circle,draw=none,minimum size=4mm, inner sep=1pt, semithick, fill=tumorange](char){\color{black}#1}}}
\newcommand{\kruggg}[1]{\tikz[baseline=(char.base)]{\node[shape=circle,draw=none,minimum size=4mm, inner sep=1pt, semithick, fill=tumgreen](char){\color{black}#1}}}
\newcommand{\kvadrat}[1]{\tikz[baseline=(char.base)]{\node[fill=white, shape=rectangle,draw,minimum size=4mm, inner sep=1pt, semithick](char){#1}}}
\newcommand{\dijamant}[1]{\tikz[baseline=(char.base)]{\node[fill=white, shape=star, star points=5,draw,minimum size=4mm, inner sep=1pt, semithick](char){#1}}}
\newcommand{\myTikzCircle}{
  \begin{tikzpicture}[baseline=-0.6ex],
    \node [circle, fill=yellow, scale=0.75, draw=black, line width = 0.1mm] {};
  \end{tikzpicture}
}

\maketitle
\thispagestyle{empty}
\pagestyle{empty}

\begin{strip}
  \vspace*{-17mm}

   \includegraphics[trim={25mm 0mm 25mm 0mm},clip,width=\textwidth]{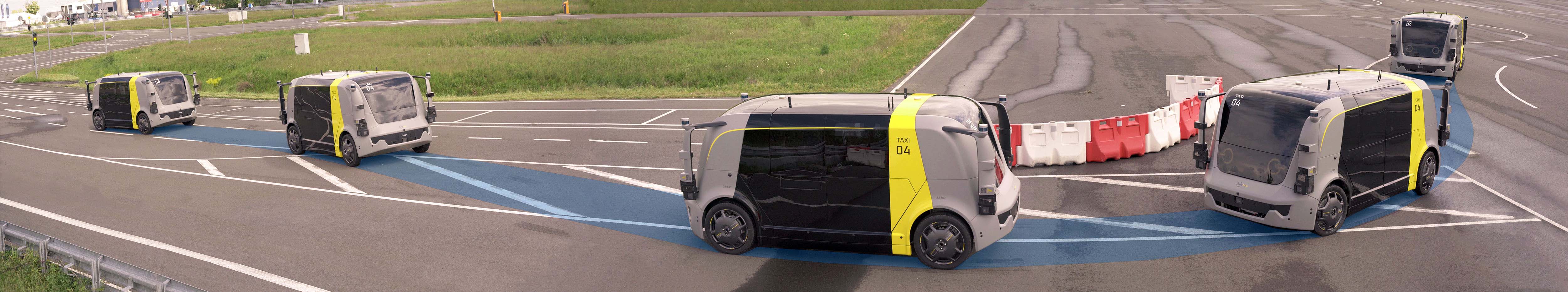}%
   \begin{tikzpicture}[overlay]
      \fill[white, draw=black, thick] (-17.0,1.2) rectangle (-16.5,1.7);
      \fill[white, draw=black, thick] (-0.75,2.8) rectangle (-0.25,3.3);
      \node at (-16.75,1.45) {B};
      \node at (-0.5,3.05) {A};
      \node at (-5.0,2.55) {\dijamant{C}};
      \node [circle, fill=yellow, scale=0.75, draw=black, line width = 0.1mm] at (-1.4,1.5) {};
      \node [circle, fill=yellow, scale=0.75, draw=black, line width = 0.1mm] at (-3.4,0.8) {};
      \node [circle, fill=yellow, scale=0.75, draw=black, line width = 0.1mm] at (-8.3,0.65) {};
      \node [circle, fill=yellow, scale=0.75, draw=black, line width = 0.1mm] at (-13.4,1.55) {};
      \node [circle, fill=yellow, scale=0.75, draw=black, line width = 0.1mm] at (-16.6,1.95) {};

      \draw (-1,1) node[] (E) {};  
      \draw (-12.5,0.8) node[label={[text=yellow]below:{Driving direction}}] (F) {};

      \draw[yellow, very thick, ->]  (E) .. controls +(-5,-1.75) and +(1.5,-0.25).. (F);
    \end{tikzpicture}
   \vspace*{-6mm}
   \captionof{figure}{A teleoperated autoTAXI vehicle maneuvering around an improvised construction site \protect\dijamant{C} on a test track in Aldenhoven, Germany. The vehicle tracks the trajectory defined by the remote operator (path visualized in blue) from the start position \protect\kvadrat{A} to the goal position \protect\kvadrat{B}, effectively solving the AV disengagement scenario. Yellow circles (\hspace{-3pt}\protect\myTikzCircle{}\hspace{-3pt}) depict approximate path waypoints specified by a remote operator using the presented teleoperation system.}
\end{strip}

\begin{abstract}

Despite the rapid technological progress, autonomous vehicles still face a wide range of complex driving situations that require human intervention. Teleoperation technology offers a versatile and effective way to address these challenges. The following work puts existing ideas into a modern context and introduces a novel technical implementation of the trajectory guidance teleoperation concept. The presented system was developed within a high-fidelity simulation environment and experimentally validated, demonstrating a realistic ride-hailing mission with prototype autonomous vehicles and onboard passengers. The results indicate that the proposed concept can be a viable alternative to the existing remote driving options, offering a promising way to enhance teleoperation technology and improve overall operation safety.
\end{abstract}

\section{Introduction}
\ac{ad} is becoming the cornerstone of future mobility. This shift from human-driven to autonomous vehicles is expected to offer a multitude of advantages, including improved efficiency, safety, and accessibility in transportation, contributing to more sustainable and efficient mobility systems. However, despite rapid technological progress, \acp{av} still face a wide array of complex and ever-changing driving situations that test their independent decision-making abilities, often even exceeding their \ac{odd}. Recently, teleoperation technology has gained prominence as a reliable alternative for \ac{ad}. It offers a versatile and effective way to address various challenges throughout the \ac{ad} operational cycle. Beyond being just a backup, teleoperation is becoming increasingly important in bringing \acp{av} to the public roads and scaling up the technology \cite{Sexton2023-wb}.

\subsection{Teleoperation as a Fallback Solution}
The primary goal of teleoperation is to support the \ac{av} operation by allowing human remote operators to take over remote control of the \ac{av} and help resolve challenging situations safely and efficiently. Over the last two decades, academia and industry have explored and demonstrated various concepts of remote interaction with AVs. These concepts are usually categorized into two main groups based on the level of human involvement, interaction type, and overall use case \cite{Majstorovic2022-px,Amador_undated-yn}.
The first category, \ac{rd}, involves teleoperation concepts where the operator takes responsibility for the driving task.
Conversely, the second group of concepts, named \ac{ra}, lets operators assist the \acp{av} by providing high-level guidance while the vehicle remains in charge of the driving task.

\subsection{Project UNICAR.agil and Research Questions}
The concept groups have different system prerequisites and can be used for different use cases.
The project UNICARagil \cite{noauthor_undated-nr} provided a platform to explore the effectiveness of \ac{rd} concepts in a real-world setting.
Funded by the German government, the project created the basis for sustainable and intelligent user-centered transportation vehicles of the future \cite{Michael_Buchholz_undated-ze}.
Four driverless vehicle prototypes were developed, each with a different use case focus (see Table \ref{tab:unicar_vehicles} below). These use cases consider close interaction between the automated driving and teleoperation modes, where the remote operator can take over the vehicle control at any point.
\vspace*{-8mm}
\begin{table}[!b]
	\caption{UNICARagil \ac{av} prototypes and their use cases}
	\label{tab:unicar_vehicles}
	\centering
	\begin{tblr}{stretch=1.2, colspec={Q[c,1.5cm]|Q[c,1.5cm]|Q[c,2.0cm]|Q[c,1.5cm]},rowspec={Q[m]Q[m]|},row{1} = {bg=whitesmoke}}
		\hline[1pt]
		\textbf{autoCARGO}  & \textbf{autoELF} & \textbf{autoSHUTTLE} & \textbf{autoTAXI} \\
		\hline[1pt]
		Delivery & Family & Ridepooling & Ride-hailing
	\end{tblr}
\end{table}

\newpage
 Additionally, vehicles used the cloud infrastructure \cite{Lampe2019-vi} to report the disengagement events and request assistance. Typical scenarios include navigating complicated traffic interactions (e.g., a policeman at an intersection waiving the right of way) or navigating construction sites, roadworks, and other complex situations not covered by the nominal \ac{odd}. Placing \acp{av} into some of these situations offered a unique opportunity to explore how to improve the current remote driving \emph{State-of-the-Art} and motivated the following research questions:
\begin{itemize}
  \item \emph{How to go beyond the \ac{dc} while still providing the operator with a high level of control?}
  \item \emph{How to go beyond the steering wheel and pedal input modality while still providing the operator with an intuitive way of interacting with the vehicle?}
  \item \emph{How to design a corresponding HMI that enables the operator to build-up situational awareness and make informed decisions?}
\end{itemize}

\subsection{Contribution and Structure}
This work introduces a novel technical implementation of the \emph{\ac{tg}} \ac{tc} \cite{Majstorovic2022-px} to enhance the current state of the remote driving technology. The presented work puts the existing ideas into a modern context and provides a technical implementation of all the essential components and their interactions. A high-fidelity simulation environment supported the development of the system. Finally, a series of driving tests with real \ac{av} prototypes under realistic disengagement events offered a way to conduct initial experimental validation of the system, effectively demonstrating advantages and disadvantages compared to the state-of-the-art. The paper is structured as follows: Section \ref{sec:relatedwork} gives an overview of the related work and motivates Section \ref{sec:approach}, which details the system strategy and architecture. The methodology for experimental validation is introduced in Section \ref{sec:validation}, with Section \ref{sec:results} providing a discussion on results and outlining future work possibilities. Finally, Section \ref{sec:conclusion} concludes the paper.

\section{Related Work}
\label{sec:relatedwork}
The idea of improving the direct control teleoperation concept has been around for decades. It was mainly motivated by the need to reduce the vulnerability of the teleoperation system to communication delays and network instabilities \cite{Georg2019-dt}. Additionally, relieving the operator from the burden of controlling every aspect of the driving task was seen as a way to reduce workload and increase safety. One popular way of achieving this is by using \ac{sc} TC \cite{Schimpe2022-up}. While it uses the same input modality as the \ac{dc}, the onboard shared controller analyzes the operator's input and modifies it to ensure safety (if necessary). In contrast to \ac{dc} and \ac{sc}, diverging from the vehicle stabilization task, the idea of \ac{tg} was introduced \cite{Kay1995-ab}. Here, the operator provides the vehicle with a trajectory to follow. The vehicle then uses its onboard localization modules and controller to track it. This responsibility reduction aimed to improve the operator's \ac{sa} and reduce the workload while simultaneously making the system more robust to network issues. Different implementations of the trajectory guidance concept with more or less details have been proposed over the years. Gnatzig \emph{et al.} \cite{Gnatzig2012-bu} suggested using a steering wheel or joystick to provide the system with shorter trajectories up to several meters long. However, the fact that the operator has to continuously provide the vehicle with new trajectory segments is a significant drawback. One way to address this issue is to introduce a mouse and keyboard as input modalities to define the trajectory that now could be much longer and more complex. Kay and Thorpe \cite{Kay1995-ab} proposed a camera-based HMI concept that allows the operator to define such trajectory by clicking on the image of the road ahead. However, this teleoperation system was never technically implemented or evaluated. On the other hand, Hoffmann \emph{et al.} \cite{Hoffmann2022-cx} used the \ac{tg} idea to develop a safety concept for the \ac{dc} mode, generating a \ac{mrm} based on operator command(s) that would be tracked in case of emergency. Finally, discussing the functionalities of a modern \ac{av}, Jatzkowski \emph{et al.} \cite{Jatzkowski2021-bx} suggested that \ac{tg} might play an essential role as one of the standard \ac{rd} teleoperation concepts, but provided no further implementation details.
This lack of technical implementation - especially in modern \ac{av} context - motivated the following work. Additionally, the availability of vehicle prototypes offered a suitable HW/SW platform \cite{Kampmann2019-eg} for experimental validation in a real-world setting.

\section{Approach and Implementation}
\label{sec:approach}
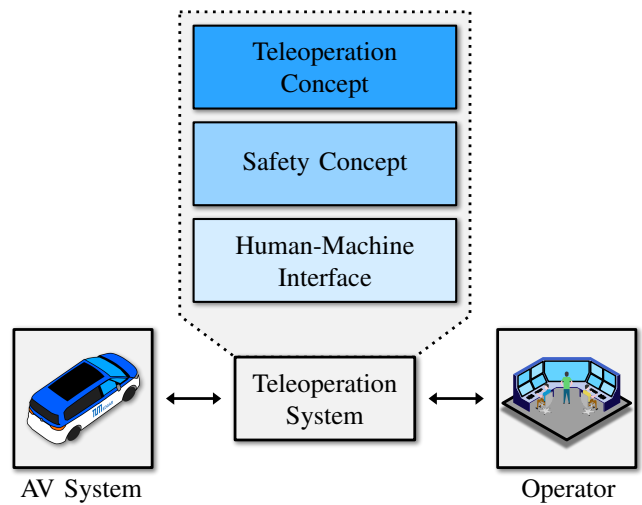
\begin{figure}[b]
    \includegraphics[trim={0mm 0mm 0mm 0mm},clip,width=\columnwidth]{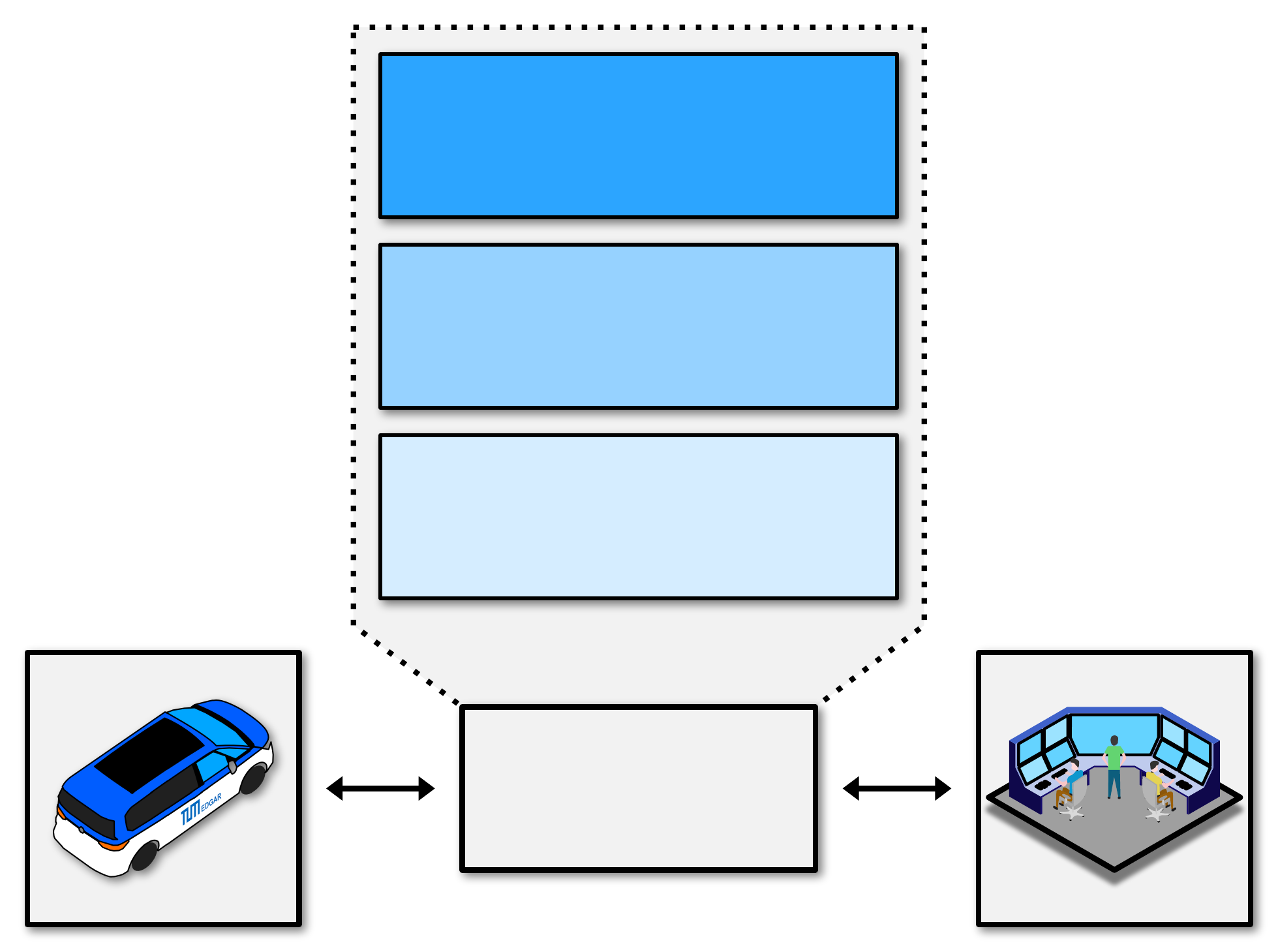}
    \begin{tikzpicture}[overlay]
      \node[align=center] at (0.125\columnwidth,0.3) {AV System};
      \node[align=center] at (0.875\columnwidth,0.3) {Operator};
      \node[align=center] at (0.5\columnwidth,1.48) {Teleoperation \\ System};
      \node[align=center] at (0.5\columnwidth,3.35) {Human-Machine \\ Interface};
      \node[align=center] at (0.5\columnwidth,4.65) {Safety Concept};
      \node[align=center] at (0.5\columnwidth,5.9) {Teleoperation \\ Concept};
    \end{tikzpicture}
    \captionof{figure}{Main components of the teleoperation system.}
    \label{Fig:tof_system}
\end{figure}

Enabling interaction between the remote operator and the \ac{av} requires a teleoperation system usually divided into three main components, as shown in Fig. \ref{Fig:tof_system}. The \emph{Teleoperation Concept} defines the technical functionalities of the system. The interface modalities and interaction features are defined by the \emph{\ac{hmi}}. Finally, the \emph{Safety Concept} describes the system's safety aspects and behavior mechanisms in safety-critical situations.
\vspace*{-4mm}
\newpage
The development of the teleoperation system used the user-centered design process defined by the ISO 9241-210 standard \cite{Iso2019-ad}. 
Table \ref{tab:requirements} shows the derived system requirements used as a basis for the technical specification of the system.

\begin{table}[!h]
	\caption{Main system requirements}
	\label{tab:requirements}
	\centering
	\begin{tblr}{stretch=1.2, colspec={Q[c,1.5cm]|Q[l,6.1cm]},rowspec={Q[m]Q[m]|Q[m]|Q[m]|Q[m]|Q[m]|Q[m]|Q[m]},row{1} = {bg=whitesmoke}}
		\hline[1pt]
		\textbf{Requirement}  & \textbf{Description} \\
		\hline[1pt]
		RQ-1\textsuperscript{*} & The vehicle shall track the trajectory and remain in charge of control signal generation and execution \\
		RQ-2\textsuperscript{*} & The teleoperation system shall be integrated within the existing ToD software architecture \\
		RQ-3\textsuperscript{*} & The teleoperation system shall support the integration with prototype vehicles \\
		RQ-4\textsuperscript{$\dagger$} & The teleoperation system shall provide an HMI that supports the build-up of \ac{sa} and vehicle interaction \\
		RQ-5\textsuperscript{$\dagger$} & The teleoperation system shall allow the remote operator to define a trajectory for the AV to follow \\
		RQ-6\textsuperscript{$\ddagger$} & The operator shall be able to trigger an MRM and stop the trajectory tracking process \\
		RQ-7\textsuperscript{$\ddagger$} & In case of emergency, a vehicle shall execute an MRM (due to network loss, collision risk, etc.) \\ \hline[1pt]
    \SetCell[r=1,c=2]{r} Requirement type: \textsuperscript{*}T. Concept, \, \textsuperscript{$\dagger$}HMI, \, \textsuperscript{$\ddagger$}Safety
    
	\end{tblr}
\end{table}
\noindent The following sections describe the main components of the developed teleoperation system in detail.

\subsection{Teleoperation Concept}
\begin{figure*}[!b]
  \vspace*{-5mm}
   \includegraphics[trim={0mm 0mm 0mm 0mm},clip,width=\textwidth]{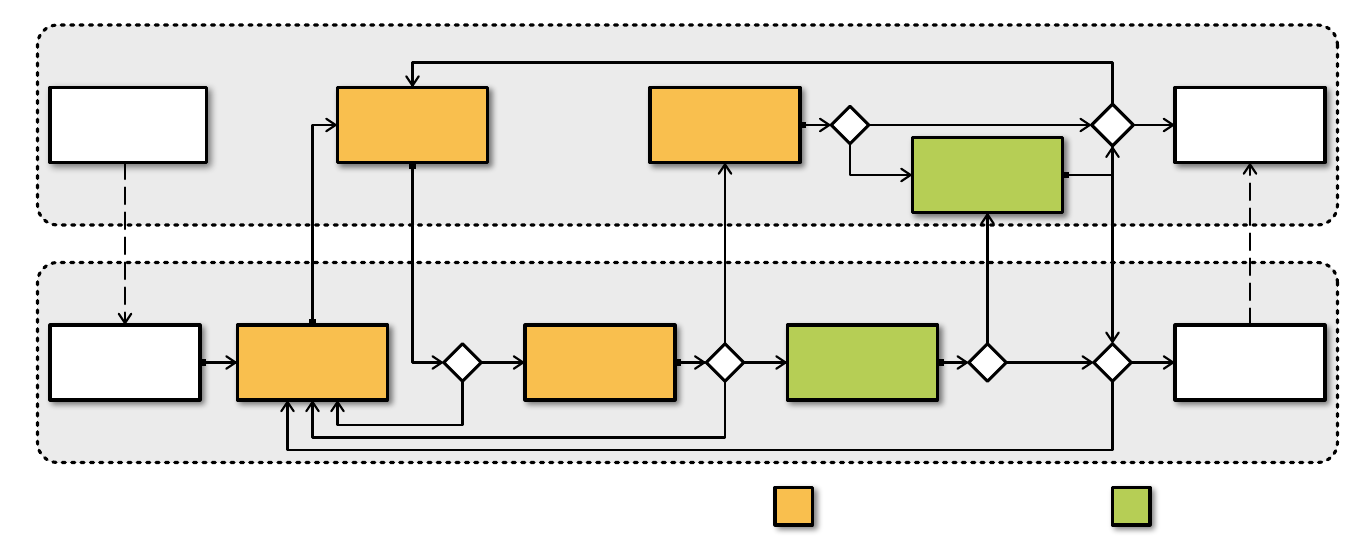}
   \begin{tikzpicture}[overlay]
    \node[align=center, rotate=90] at (0.032\columnwidth,6.00) {AV System};
    \node[align=center, rotate=90] at (0.032\columnwidth,2.9) {Remote Operator};
    \node[align=center] at (0.195\columnwidth,6.00) {Teleoperation \\ Request};
    \node[align=center] at (0.195\columnwidth,2.9) {Takeover};
    \node[align=center] at (0.480\columnwidth,2.9) {Trajectory \\ Creation};
    \node[align=center] at (0.630\columnwidth,6.00) {Trajectory \\ Check};
    \node[align=center] at (0.920\columnwidth,2.87) {Trajectory \\ Approval};
    \node[align=center] at (1.315\columnwidth,2.87) {Monitoring};
    \node[align=center] at (1.105\columnwidth,6.00) {Trajectory \\ Tracking};
    \node[align=center] at (1.505\columnwidth,5.35) {Emergency \\ Stop};
    \node[align=center] at (1.905\columnwidth,6.00) {Automated \\ Operation};
    \node[align=center] at (1.905\columnwidth,2.9) {Handover};
    \node[align=center] at (1.450\columnwidth,0.92) {Teleoperation Concept};
    \node[align=center] at (1.905\columnwidth,0.92) {Safety Concept};

    \node [anchor=west, color=black] (note) at (3.35,4.45) {$r_{\textrm{ref}}$};
    \node [anchor=west, color=black] (note) at (5.4,2.4) {\small\krugg{1}};
    \node [anchor=west, color=black] (note) at (6.2,3.2) {\small\krugg{2}};
    \node [anchor=west, color=black] (note) at (8.9,2.2) {\small\krugg{3}};
    \node [anchor=west, color=black] (note) at (8.9,5.05) {\small\krugg{4}};
    \node [anchor=west, color=black] (note) at (9.6,3.2) {\small\kruggg{8}};
    \node [anchor=west, color=black] (note) at (11.2,5.05) {\footnotesize\kruggg{10}};
    \node [anchor=west, color=black] (note) at (11.2,6.35) {\small\krugg{5}};
    \node [anchor=west, color=black] (note) at (14.05,6.55) {\small\krugg{6}};
    \node [anchor=west, color=black] (note) at (12.35,3.9) {\small\kruggg{9}};
    \node [anchor=west, color=black] (note) at (13.1,3.2) {\footnotesize\kruggg{11}};
    \node [anchor=west, color=black] (note) at (14.0,5.05) {\footnotesize\kruggg{12}};
    \node [anchor=west, color=black] (note) at (14.0,2.05) {\small\krugg{7}};
  \end{tikzpicture}
  \vspace*{-5mm}
   \captionof{figure}{Trajectory guidance operation logic showcasing the teleoperation concept and safety mechanism elements.}
   \label{Fig:tg_logic}
\end{figure*}
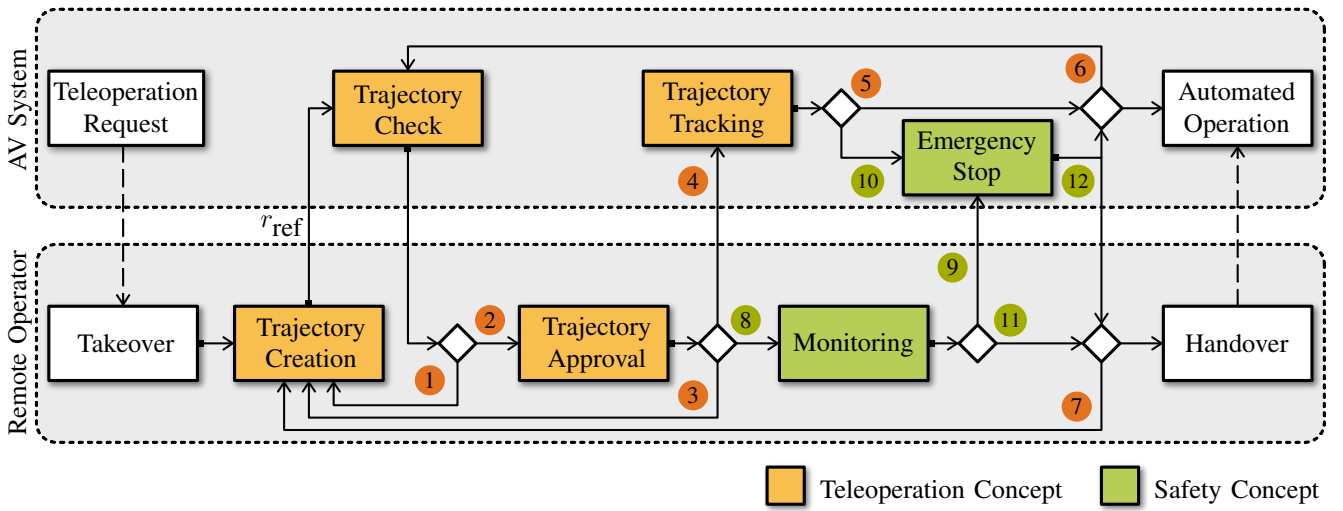

The overall operational logic of the trajectory guidance teleoperation system can be seen in Fig. \ref{Fig:tg_logic}. Main concept phases (shown in orange) describe the interaction between the remote operator and the vehicle. Additionally, two main safety mechanisms (shown in green) depict the safety concept that ensures safe operation.
As the \ac{av} system disengages its operation, it sends a teleoperation request\footnote{In a realistic production setting, the \ac{av} reports the disengagement event using cloud infrastructure (possibly as a digital twin) to the operation center, which assigns a human remote operator to assist the vehicle in need.}
to the remote operator. An established connection between the operator and the vehicle marks the moment of the \emph{Takeover}.

\subsubsection{Trajectory Creation}
\begin{figure}[H]
  \centering
  \includegraphics[trim={0mm 27mm 0mm 8mm},clip,width=\columnwidth]{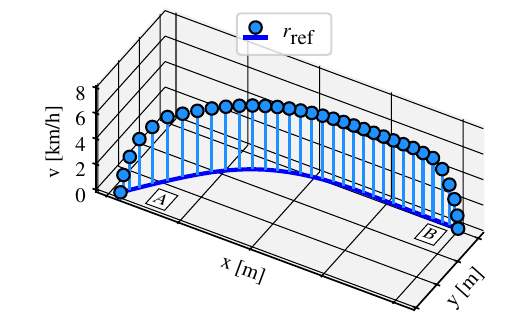}
  \captionof{figure}{Reference trajectory, $r_{\textrm{ref}}$ produced during the trajectory creation phase. The vehicle is expected to track this trajectory.}
  \label{Fig:trajectory_creation}
\end{figure}

In the first phase, the operator creates a trajectory for the vehicle to follow, defined by a series of path points with a corresponding velocity profile. The trajectory is created as follows: The operator uses the sensor data to assess the situation, makes a decision about what the vehicle should do, and uses the \ac{hmi} to define path waypoints up to the goal position - the number of waypoints depends on the path length. Once specified, a smooth path with equidistant points is generated using cubic $C^2$ spline interpolation \cite{Kluge2021-iq}. Finally, this path is assigned a trapezoidal velocity profile \cite{Lynch_undated-cr} generated for each path point while considering the path topology (curvature) and maximum allowed velocity/acceleration/jerk constraints to ensure smooth driving behavior. Fig. \ref{Fig:trajectory_creation} shows the result of this procedure.

\subsubsection{Trajectory Check}
Once created, the trajectory is sent to the vehicle, which verifies the data integrity and performs final feasibility and safety checks. If the trajectory check fails {\small\krugg{1}}, the process returns to the previous stage, where the operator has to create a new trajectory. Otherwise, the process continues {\small\krugg{2}} to the next phase. In both cases, the HMI informs the operator about the status of the trajectory check.

\subsubsection{Trajectory Approval}
In this phase, the operator does the final review of the trajectory. Suppose the traffic situation has changed to the point where the previous decision is not satisfying anymore. In that case, the operator can decide to reject it {\small\krugg{3}} and return to the \emph{Trajectory Creation} phase to create a new one. Otherwise, the operator approves the trajectory {\small\krugg{4}}, which triggers the next phase.

\subsubsection{Trajectory Tracking}
Once the vehicle receives the approval, it does the final processing of the trajectory to fit the required input specification. This depends on the controller type (e.g., see Homolla \emph{et al.} \cite{Homolla2022-jq}), but it is necessary to ensure proper communication with the controller. The vehicle then starts to track the trajectory {\small\krugg{5}} while streaming the vehicle data to the operator. The trajectory tracking runs until the vehicle reaches the specified end of the path, where it stops and transits back to the \emph{Trajectory Check} state {\small\krugg{6}} awaiting the subsequent trajectory from the operator {\small\krugg{7}}. If the vehicle has reached the goal position, the operator can end the teleoperation session, the \ac{av} will return to the \emph{Automated Operation} state and continue its original autonomous mission.

\subsection{Safety Concept}
\subsubsection{Monitoring}
As in all remote driving concepts, the operator takes full responsibility for the safety of the vehicle's behavior. Consequently, as the \ac{av} starts tracking the trajectory, the operator must actively monitor the vehicle {\small\kruggg{8}} and intervene if necessary. The operator can trigger this intervention at any time and will transit the system to the \emph{Emergency Stop} state {\small\kruggg{9}}. Additionally, the vehicle will also traverse into the \emph{Emergency Stop} state automatically if it detects a safety-critical situation {\footnotesize\kruggg{10}} such as complete network loss, collision risk, etc.
\subsubsection{Emergency Stop}
\begin{figure}[!t]
  \centering
  \includegraphics[trim={0mm 27mm 0mm 8mm},clip,width=\columnwidth]{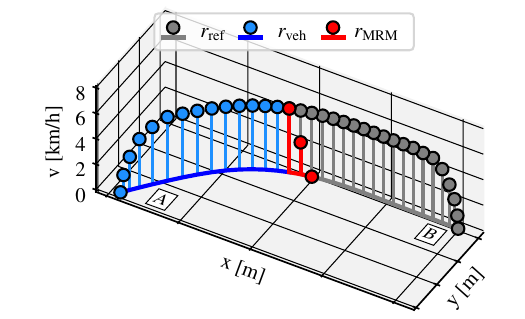}
  \captionof{figure}{Simulated execution of the \ac{mrm} during the trajectory tracking phase.}
  \label{Fig:MRM}
  \vspace*{-6mm}
\end{figure}

In any case, the vehicle will trigger an \ac{mrm} that ensures the safety of both ego-vehicle and other road participants, inform the operator about the event, and proceed to wait for the next trajectory {\footnotesize\kruggg{12}} or session end. Fig. \ref{Fig:MRM} shows the maneuver execution process amid the trajectory tracking phase. Each time step of the trajectory tracking triggers a new maneuver generation with an adapted path and velocity profile.

\subsection{Human-Machine Interface (HMI)}
\begin{figure}[!b]
  \vspace*{-5mm}
  \captionsetup[subfigure]{aboveskip=-4pt,belowskip=8pt}
  \begin{subfigure}{\columnwidth}
    \centering
    \includegraphics[trim={20mm 40mm 20mm 20mm},clip,width=1\columnwidth]{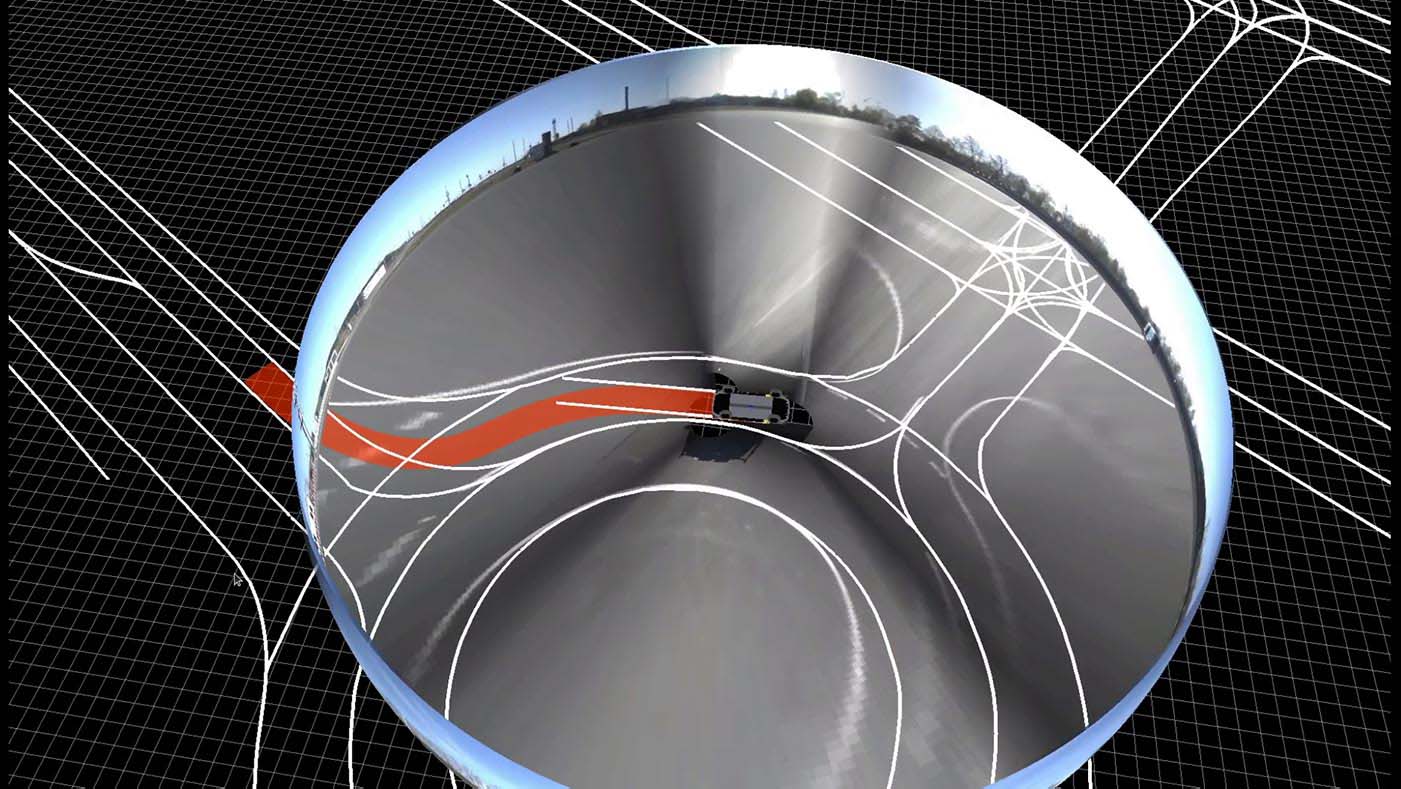}
    \begin{tikzpicture}[overlay,shift={(3.75,1)}]
      \node[below right] at (-3.8,1.0) {\krug{A}};
      \node[] at (-4.9,3.0) {\krug{B}};
      \node[] at (-7.5,0.5) {\krug{C}};
      \node[] at (-4.9,1.5) {\krug{D}};
    \end{tikzpicture}
    \subcaption{HMI v1: TG teleoperation GUI based on ToD SW Stack \cite{Schimpe2022-ig}}
  \end{subfigure}
  \\
  \begin{subfigure}{\columnwidth}
    \centering
    \includegraphics[trim={0mm 0mm 0mm 10mm},clip,width=1\columnwidth]{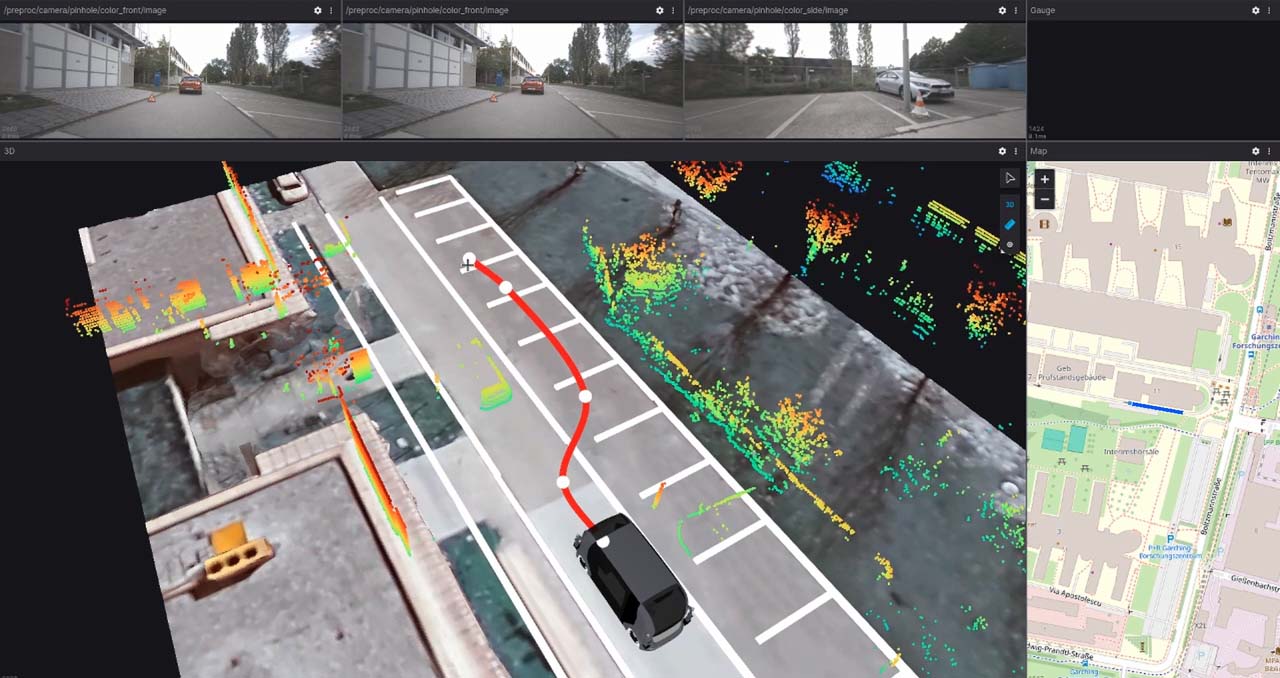}
    \begin{tikzpicture}[overlay,shift={(4.2,0.5)}]
    \node[] at (-4.8,0.4) {\krug{A}};
    \node[] at (-4.9,3.75) {\krug{B}};
    \node[] at (-3.3,0.5) {\krug{C}};
    \node[] at (-4.6,2.0) {\krug{D}};
    \node[] at (-2.8,3.0) {\krug{E}};
    \node[] at (-6.3,0.5) {\krug{F}};
    \node[] at (-0.9,1.0) {\krug{G}};
  \end{tikzpicture}
    \subcaption{HMI v2: TG teleoperation GUI based on Foxglove \cite{noauthor_undated-xw}}
  \end{subfigure}
\caption{Developed HMI concepts for the trajectory guidance teleoperation system.}
\label{Fig:HMI}
\end{figure}
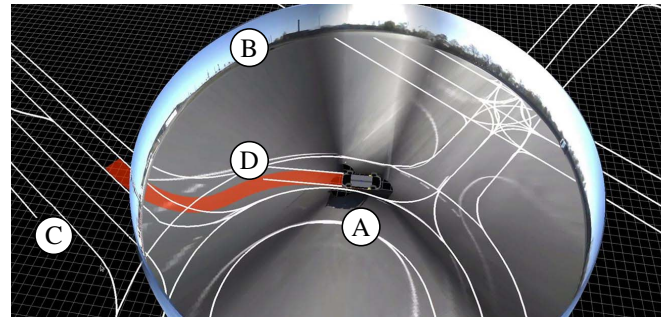
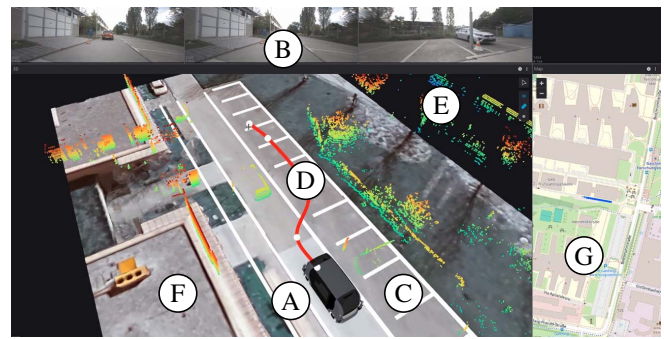

\ac{hmi} is the main interface between the operator and the vehicle. It provides the operator with the necessary information about the vehicle and its surroundings and plays a crucial role in the operator's situational awareness. Additionally, it allows the operator to interact with the vehicle. It can be divided into two main components: \emph{Input} and \emph{Output}. Typical input devices include a steering wheel, pedals, joystick, mouse, keyboard, etc. The output part is usually realized using a combination of different graphical user interface elements displayed on one or more screens in different constellations. The (optimal) \ac{hmi} design depends on many factors, including the use case, teleoperation concept, vehicle platform, etc., and as such, is a complex topic that addresses many different aspects \cite{Tener2022-yv}. Fig. \ref{Fig:HMI} shows the developed \ac{hmi} concepts for the presented trajectory guidance teleoperation system that meets the previously defined requirements, accompanied by Table \ref{tab:hmi} that describes the main display elements and their corresponding data sources. Both concepts use the same input and output devices: a mouse \& keyboard and a display. The current industry standard \cite{Cruise2021-en,Zoox2020-ae} inspired the design.

\newpage
\begin{table}[]
	\caption{Human-Machine Interface - display elements and corresponding data sources.}
	\label{tab:hmi}
	\centering
	\begin{tblr}{stretch=1.2, colspec={Q[c,1.8cm]Q[c,1.0cm]Q[c,1.8cm]Q[c,1.8cm]},rowspec={Q[m]Q[m]|Q[m]|Q[m]|Q[m]|Q[m]|Q[m]|Q[m]},row{1} = {bg=whitesmoke}}
		\hline[1pt]
		\textbf{Element}  & \textbf{Position} & \textbf{HMI v1} & \textbf{HMI v2} \\
		\hline[1pt]
		Vehicle Data\textsuperscript{*} \& 3D Model\textsuperscript{$\dagger$} & \krug{A} & {\usym{1F5F8}} & {\usym{1F5F8}} \\
		Camera Streams\textsuperscript{*} & \krug{B} & {\usym{1F5F8} \\ 3D Sphere\cite{Schimpe2022-ig}} & {\usym{1F5F8} \\ 2D Rectangles} \\
		HD-Map\textsuperscript{$\dagger$} & \krug{C} & {\usym{1F5F8} \\ Lanelet2\cite{Poggenhans2018-vm}} & {\usym{1F5F8} \\ Lanelet2\cite{Poggenhans2018-vm}} \\
		Trajectory Path\textsuperscript{$\dagger$} & \krug{D} & \usym{1F5F8} & \usym{1F5F8} \\
		LiDAR Pointcloud\textsuperscript{*} & \krug{E} & \usym{1F5F4} & {\usym{1F5F8} \\ Obstacles-only } \\
		Satellite Image\textsuperscript{$\dagger$} & \krug{F} & \usym{1F5F4} & {\usym{1F5F8} \\ Google Earth} \\
    Geographic Map\textsuperscript{*} & \krug{G} & \usym{1F5F4} & {\usym{1F5F8} \\ OpenStreetMap} \\
		\hline[1pt]
		\SetCell[r=1,c=4]{r} Data source end: \textsuperscript{*}Vehicle, \, \textsuperscript{$\dagger$}Operator
	\end{tblr}
  \vspace*{-5mm}
\end{table}

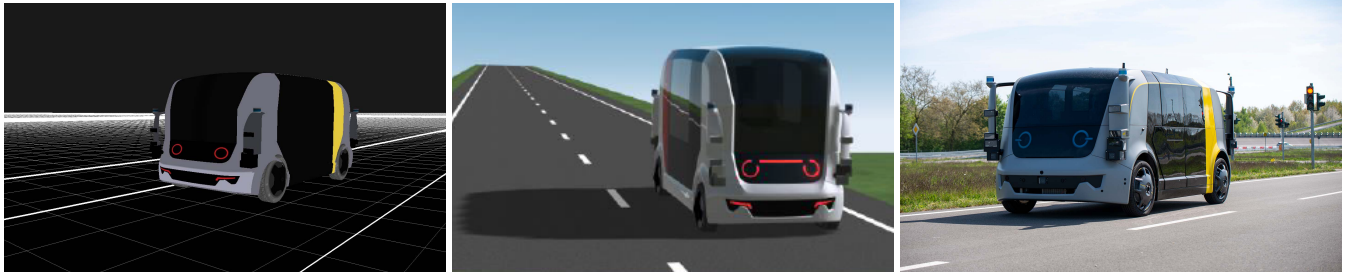
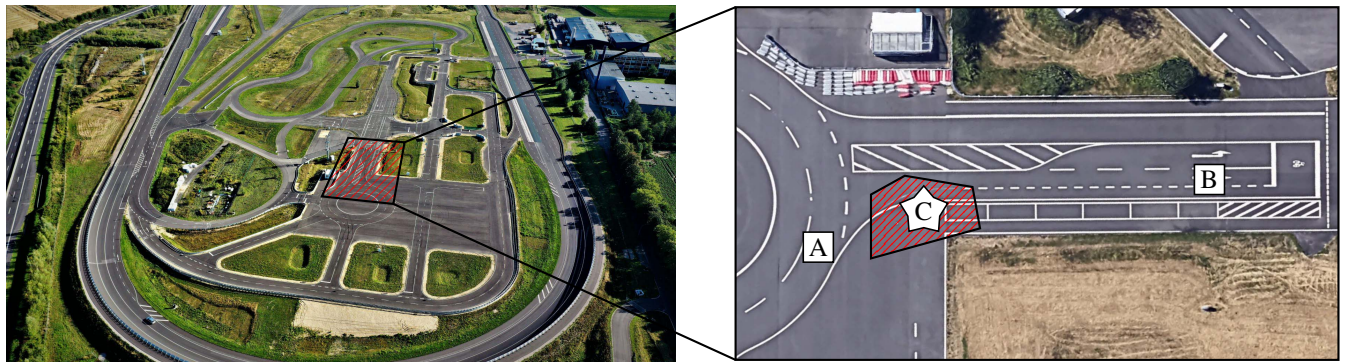
\begin{figure*}[!b]
  \vspace*{-6mm}
  \captionsetup{belowskip=6pt, aboveskip=0pt} 
  \begin{subfigure}{0.33\textwidth}
    \includegraphics[trim={150mm 40mm 0mm 25mm},clip,width=\textwidth]{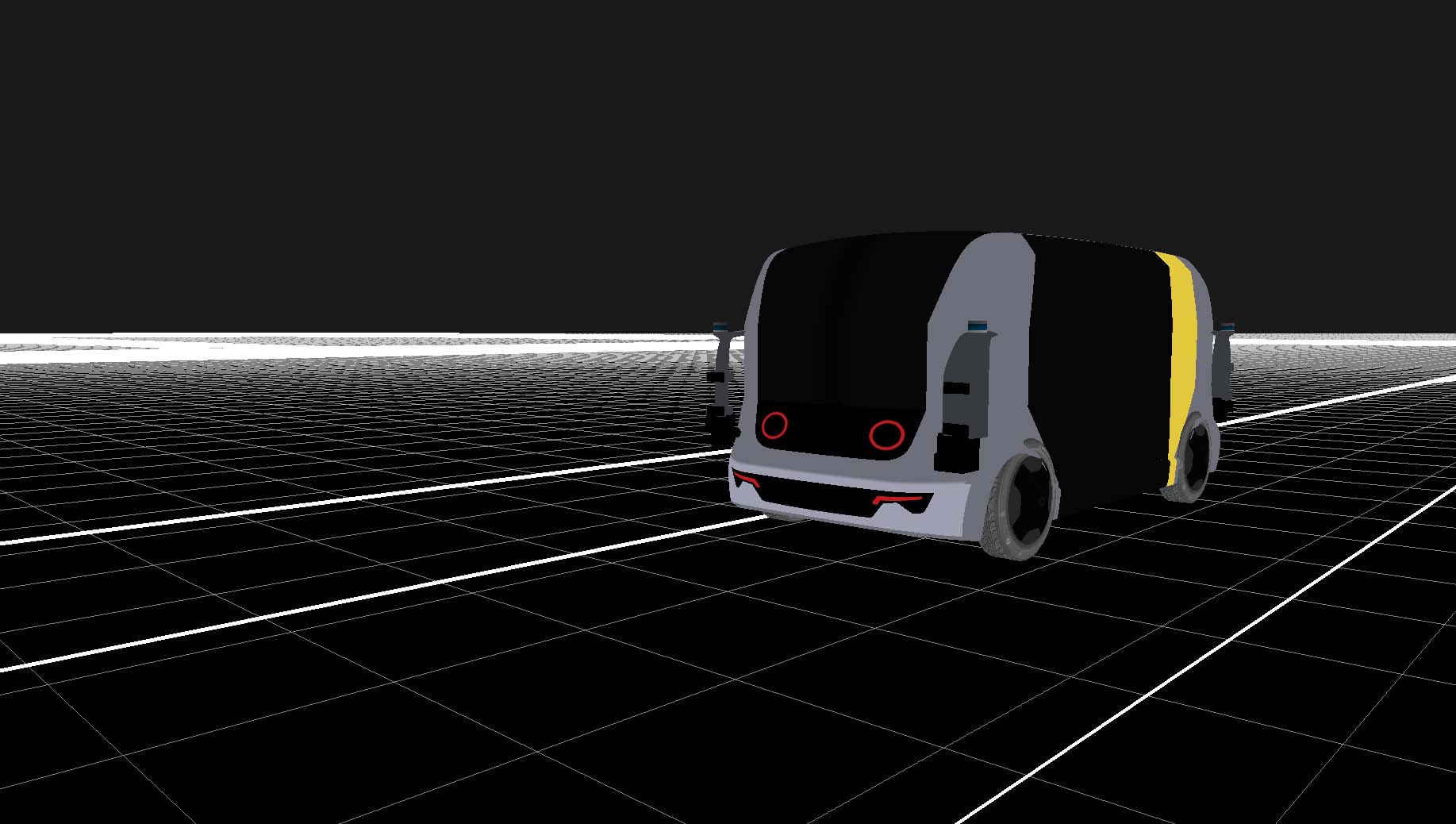}
    \subcaption{Base simulation: ToD SW \cite{Schimpe2022-ig}}
    \label{fig:tod_sw}%
  \end{subfigure}
  \begin{subfigure}{0.33\textwidth}
    \includegraphics[trim={10mm 6mm 0mm 13mm},clip,width=\textwidth]{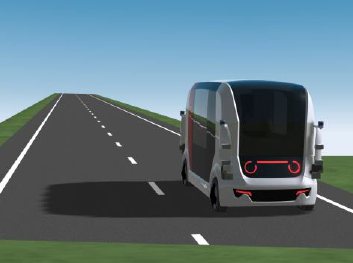}
    \subcaption{High-fidelity simulation: CarMaker \cite{noauthor_undated-yo}}
    \label{fig:carmaker}%
  \end{subfigure}
  \begin{subfigure}{0.33\textwidth}
    \includegraphics[trim={0mm 5mm 0mm 32mm},clip,width=\textwidth]{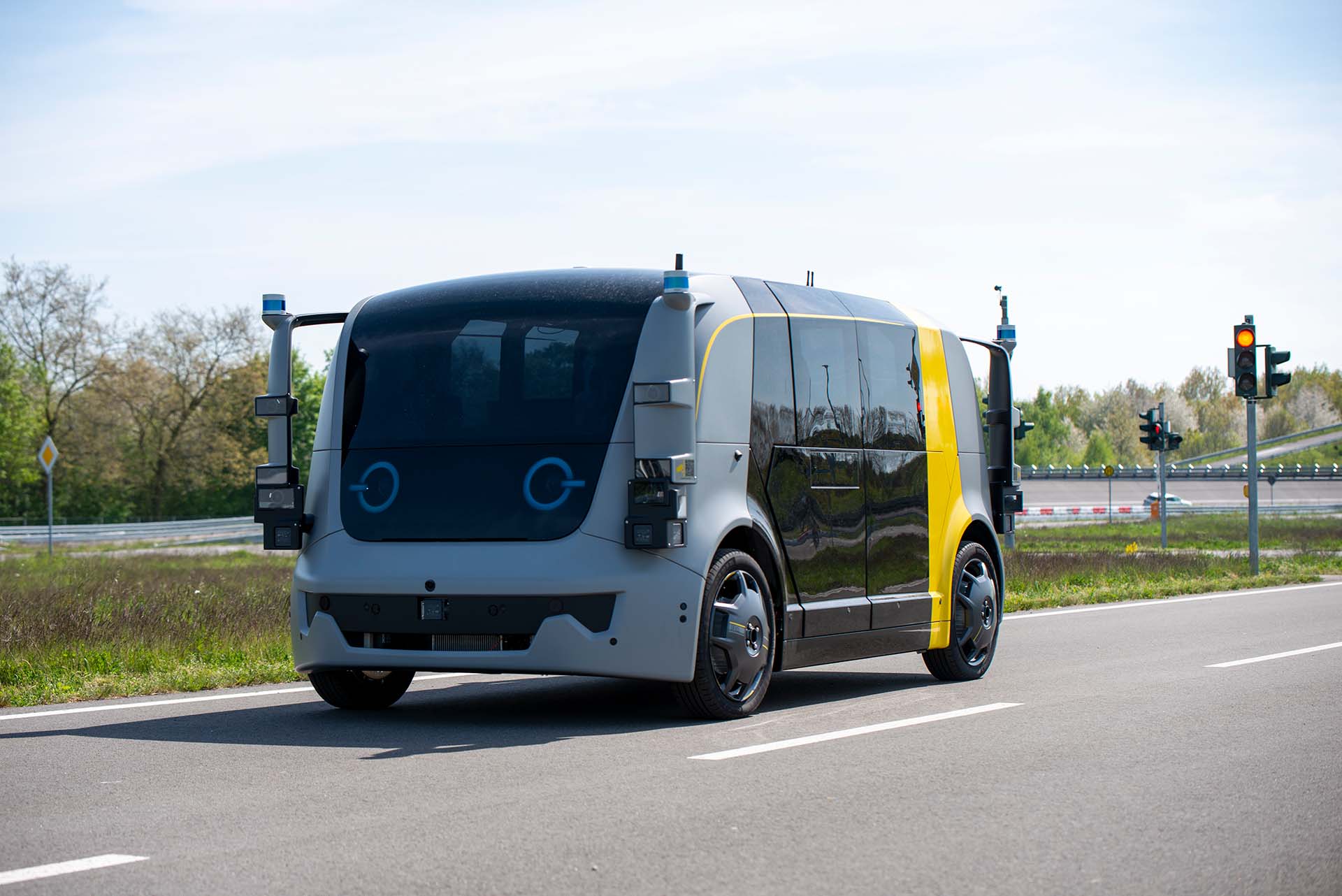}
    \subcaption{Experimental tests: \ac{av} prototypes \cite{noauthor_undated-nr}}
    \label{fig:atc}%
  \end{subfigure}
  \begin{subfigure}{1\textwidth}
    \centering
    \begin{tikzpicture}[remember picture]
      \node[anchor=south west,inner sep=0] (mainImage) at (0,0) {\includegraphics[trim={0mm 10mm 0mm 10mm},clip,width=.5\textwidth]{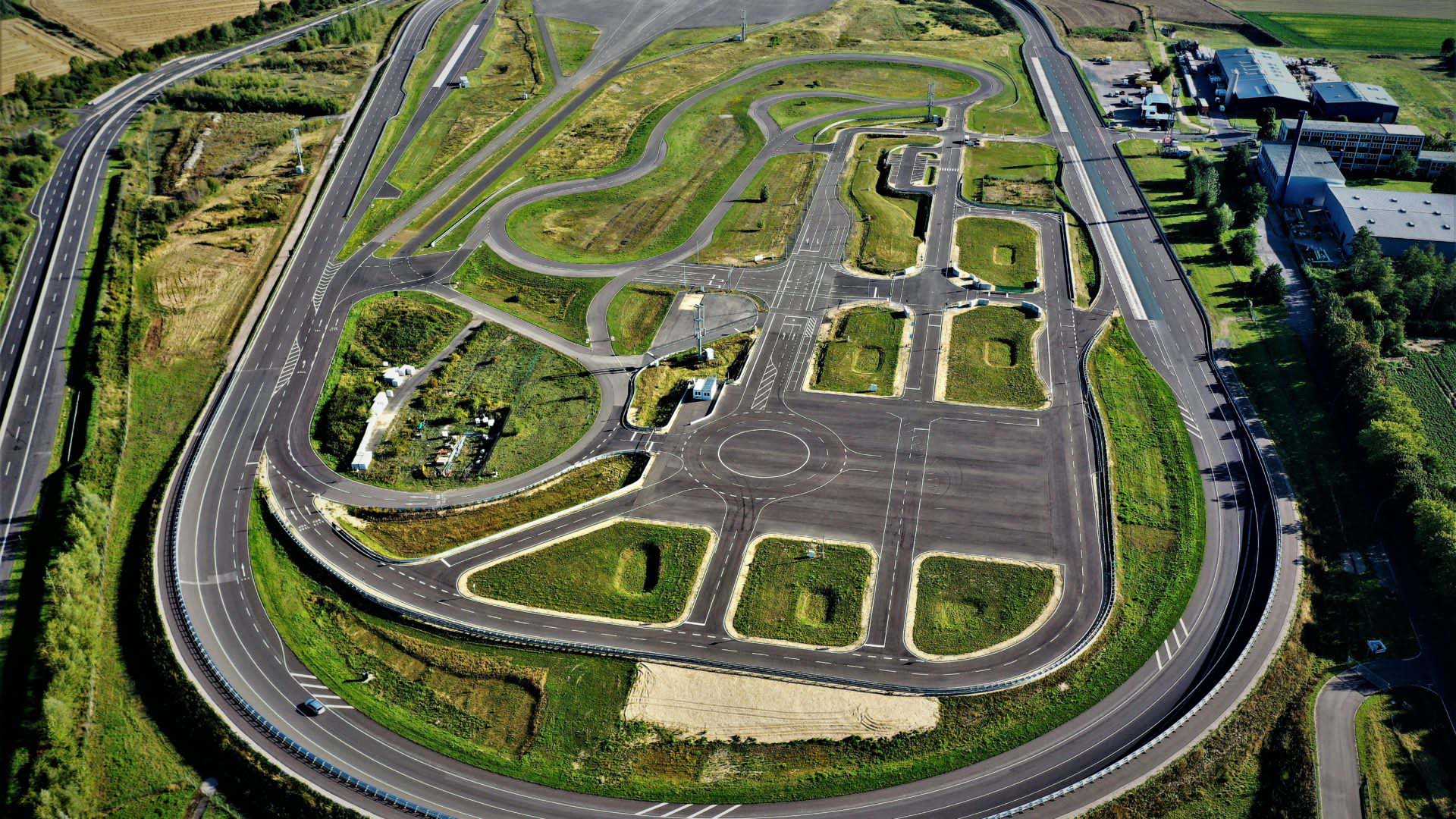}};
      \coordinate (rectCenter) at ($(mainImage.south west) + (4.2,2.2)$);
      
      \coordinate (A) at (rectCenter);
      \coordinate (B) at ($(A) + (0.35,0.78)$); %
      \coordinate (C) at ($(A) + (1.10,0.72)$); %
      \coordinate (D) at ($(A) + (0.95,-0.1)$); %
      \filldraw[black, pattern=north east lines, pattern color=red, fill opacity=0.7, thick] (A) -- (B) -- (C) -- (D) -- cycle;
  
      \node[anchor=south west,inner sep=0] (zoomedImage) at (9.7,0.025) {\includegraphics[width=.45\textwidth]{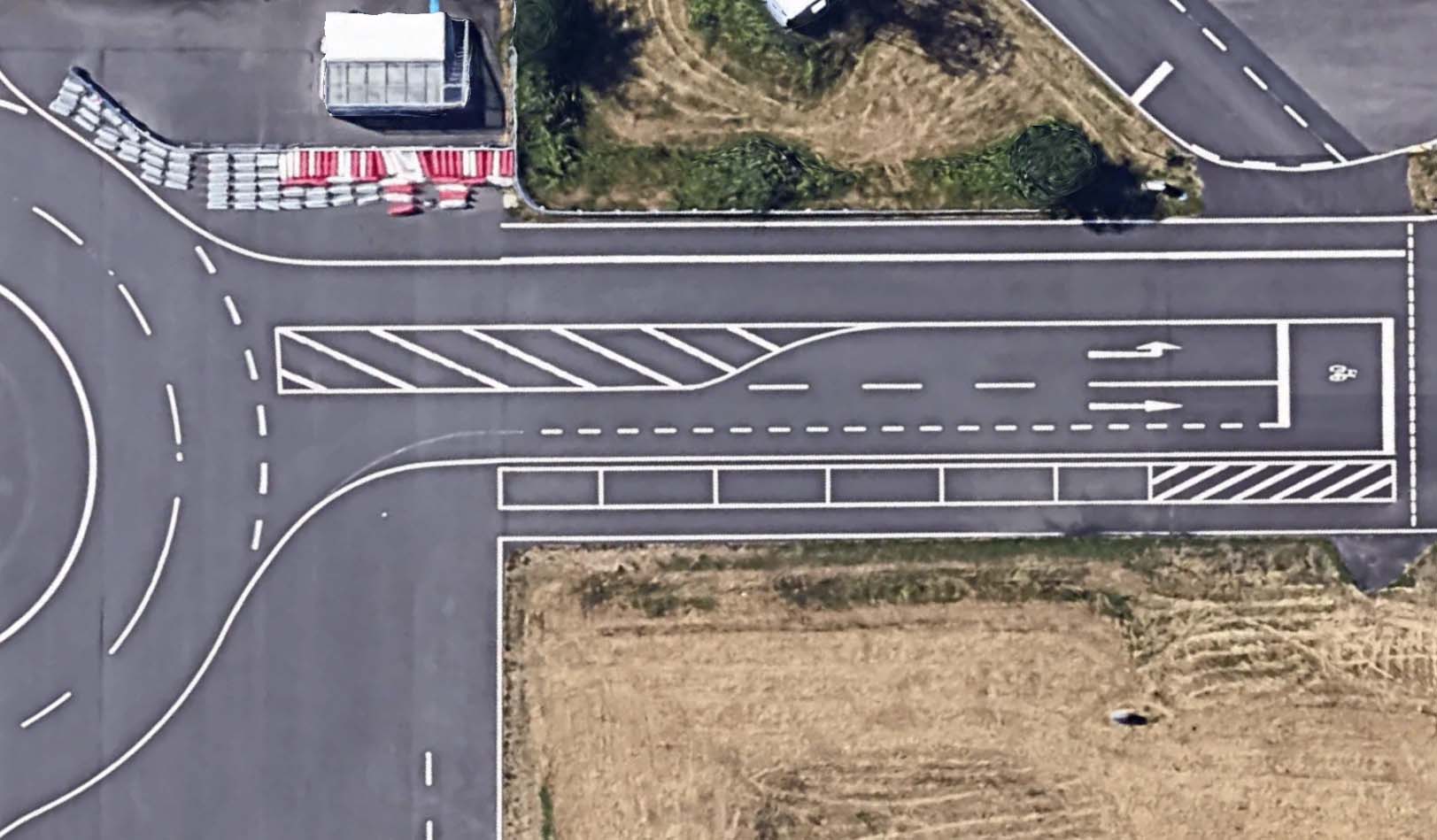}};
      \draw[thick, black] (zoomedImage.south west) rectangle (zoomedImage.north east);
  
      \draw[thick, black] (D) -- (zoomedImage.south west);
      \draw[thick, black] (C) -- (zoomedImage.north west);

      \begin{scope}[shift={(zoomedImage.south west)}]

        \coordinate (O) at (1.8cm,2.2cm);
        \coordinate (A) at (1.8cm,1.35cm);
        \coordinate (B) at (3.25cm,1.75cm);
        \coordinate (C) at (3.15cm,2.3cm);
        \coordinate (D) at (2.25cm,2.45cm);
        \filldraw[pattern=north east lines, pattern color=red, thick] (O) -- (A) -- (B) -- (C) -- (D) -- cycle;

        \coordinate (E) at (2.05cm,2.45cm);
        \node at (1.1,1.5) {\kvadrat{A}};
        \node at (6.3,2.4) {\kvadrat{B}};
        \node[below right] at (E) {\dijamant{C}};

      \end{scope}
    \end{tikzpicture}
    \vspace*{-4mm}
    \subcaption{Aldenhoven Testing Center (ATC) test track used for the showcase of the autonomous ride-hailing service at the live event (left). The zoomed area shows the construction site scenario used to demonstrate and assess the teleoperation concepts (right).}
    \label{fig:zoomedAreaConnections}
  \end{subfigure}
  \caption{Development and validation process of the presented trajectory guidance teleoperation system. A video of the ride-hailing mission demonstrated at the ATC test track is available at \url{https://youtu.be/q_jJlxmL1vo}.}
  \label{Fig:implementation}
\end{figure*}

A 15-month-long three-step development process was used to implement the presented teleoperation system, as shown in Fig. \ref{Fig:implementation}. First, the system was developed within the ToD software stack used as the main software platform (\ref{fig:tod_sw}), offering necessary teleoperation functionalities. The system was then coupled with the high-fidelity CarMaker \cite{noauthor_undated-yo} simulation software that offered a realistic simulation of the vehicle dynamics and its surroundings (\ref{fig:carmaker}). Finally, the system was experimentally tested and improved in a series of driving tests with real \ac{av} prototypes (\ref{fig:atc}).

\newpage
\section{Experimental Validation}
\label{sec:validation}
The final experimental validation of the presented teleoperation system took place at the Aldenhoven Testing Center (ATC) \cite{noauthor_undated-st} test track near Aachen, Germany (Fig. \ref{fig:zoomedAreaConnections}-left). As part of the live demonstration event, a ride-hailing service with autoTAXI vehicle was presented to the public. Interested visitors could experience the vehicle's driving capabilities as passengers on a realistic ride-hailing mission. The vehicle drove autonomously on a predefined 1.1 km-long circular route, traveling through a recreated urban environment with traffic lights, crosswalks, etc. The route had two challenging scenarios to demonstrate the capabilities of the teleoperation system: a blocked road and a construction site, each forcing the \ac{av} to disengage the autonomous operation and request human assistance. Once disengaged, the vehicle stopped, informed the operator over the cloud about the problem, and asked for a takeover. The operator in the control center (located in a separate building at the test track) had the necessary infrastructure with access to a private LTE network to take over the vehicle control. To assess the \ac{tg} system performance, the experienced operator had to teleoperate the \ac{av} and solve the construction site scenario (Fig. \ref{fig:zoomedAreaConnections}-right) using HMI v1 with both direct control (steering wheel \& pedals) and trajectory guidance (mouse \& keyboard). Each run started at the exact start location \kvadrat{A} and had an identical goal position \kvadrat{B}. During the whole-day event, the operator solved the scenario fourteen times in the following order: $7$x DC, $7$x TG. All rides had passengers onboard and were part of the autonomous mission.

\newpage
\section{Results and Discussion}
\label{sec:results}

\begin{figure*}[!b]
  \vspace*{-8mm}
  \captionsetup{belowskip=6pt, aboveskip=0pt}
  \begin{subfigure}{0.5\textwidth}
  \centering
  \includegraphics[trim={0mm 5mm 0mm 0mm},clip,width=\columnwidth]{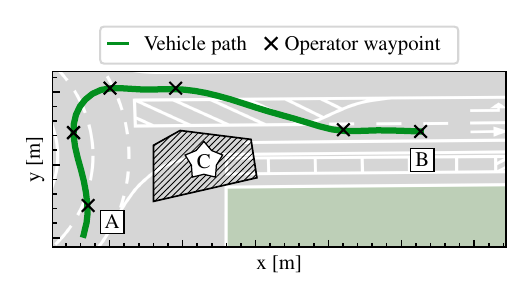}
  \subcaption{TG: Single run driving path (\#7)}
  \label{Fig:results_a}
  \end{subfigure}
  \begin{subfigure}{0.5\textwidth}
    \centering
    \includegraphics[trim={0mm 5mm 0mm 0mm},clip,width=\columnwidth]{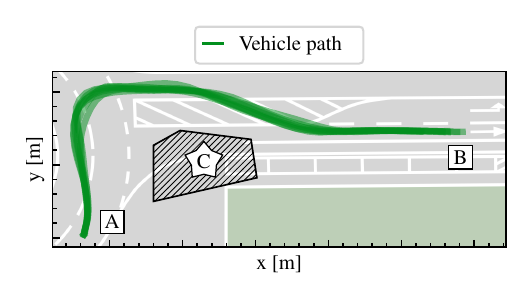}
    \subcaption{TG: All driven paths}
    \label{Fig:results_b}
    \end{subfigure}
  \\
    \begin{subfigure}{0.5\textwidth}
      \centering
      \includegraphics[trim={6mm 28mm 2mm 8mm},clip,width=\columnwidth]{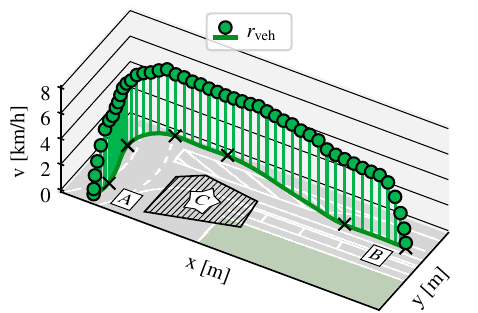}
      \subcaption{TG: Single run trajectory with path and velocity profile (\#7)}
      \label{Fig:results_c}
    \end{subfigure}
    \begin{subfigure}{0.5\textwidth}
      \centering
      \includegraphics[trim={6mm 28mm 2mm 8mm},clip,width=\columnwidth]{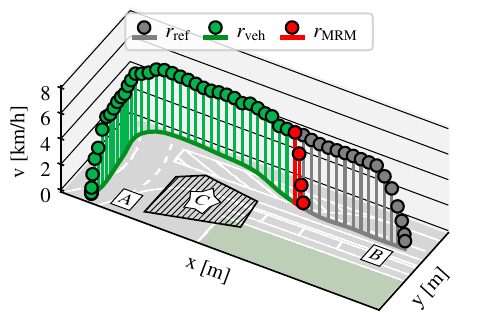}
      \subcaption{TG: Emergency MRM event triggered by the operator}
      \label{Fig:results_d}
      \end{subfigure}
  \begin{subfigure}{0.5\textwidth}
    \centering
    \includegraphics[trim={0mm 0mm 0mm 4mm},clip,width=\columnwidth]{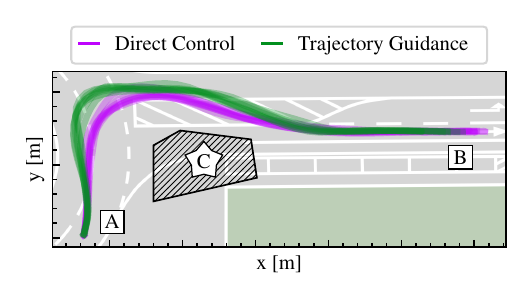}
    \subcaption{DC vs. TG: All driven paths}
    \label{Fig:results_e}
    \end{subfigure}
    \begin{subfigure}{0.5\textwidth}
      \centering
      \includegraphics[trim={0mm 2mm 0mm 6mm},clip,width=\columnwidth]{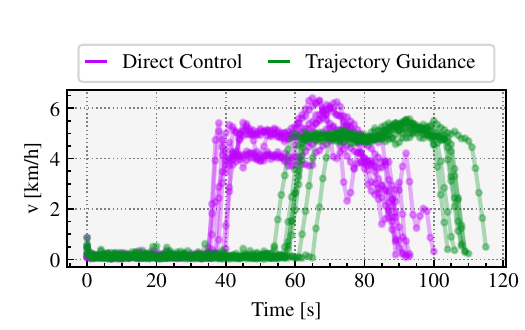}
      \subcaption{DC vs. TG: All velocity profiles}
      \label{Fig:results_f}
      \end{subfigure}
      
\caption{Experimental results obtained with the autoTAXI vehicle.}
\label{Fig:results}
\end{figure*}
Evaluating the teleoperation system in a realistic environment with the onboard passengers was done for the first time. The experiments showed that the operator could successfully solve the construction site scenario with both teleoperation concepts and guide the vehicle safely and comfortably. The main metric used to assess the performance was the total teleoperation time, defined as the time from the vehicle's disengagement to the moment the teleoperated vehicle reached the goal position. Additionally, the velocity profiles were analyzed to assess the driving behavior during each \ac{tc} operation.
A sample TG driving performance is shown in Fig. \ref{Fig:results_a}, where the crosses mark the waypoints defined by the operator. This path had the velocity profile shown in Fig. \ref{Fig:results_c} with the maximum velocity set to 5 km/h. For the Start-Goal line-of-sight distance of roughly 60 meters, the operator needed to set approximately 6-7 waypoints during the path planning phase. The topology of all 7 TG runs can be seen in Fig. \ref{Fig:results_b}. Emergency MRM was not triggered during the event, but it was tested during the late development phase and can be seen in Fig. \ref{Fig:results_d}. The operator triggered the MRM by pressing the emergency stop button on the HMI, which caused the vehicle to execute a predefined safety maneuver (with a $\sim$70 ms latency) and stop the operation (shown in red). The same system behavior was validated for the case where the vehicle triggers the MRM due to network connection loss (with an 80 ms loss detection threshold).

Adding DC runs to the topology graph puts the TG performance into perspective (Fig. \ref{Fig:results_e}). Two main differences between the concepts can be observed. First, the TG paths are much more conservative than the DC paths, keeping a safer distance from the obstacles, thus producing longer traveling paths and potentially inflicting a higher total teleoperation time. This behavior can be explained by the fact that the operator had to plan the path in advance with a limited perception of the more distant environment. Obviously, the clever use of HMI plays a crucial role. Second, the DC produced much smoother paths with greater curvature radii. This is a direct consequence of the operator having direct control over the vehicle and using the steering wheel to produce sensitive driving behavior. The TG paths, on the other hand, were generated using the predefined waypoints, which converted into a less intuitive understanding of the resulting driving style. The onboard safety driver did not report unpleasant driving behavior during any run.

Additionally, the velocity profiles of all runs can be seen in Fig. \ref{Fig:results_f}. The DC averaged at $\bar{v}_{dc}=4.09$ km/h, while TG averaged at $\bar{v}_{tg}=4.32$ km/h, which is interesting since the operator had an upper limit at $36$ km/h, and yet, rarely decided to go beyond $5$ km/h. As expected, powered by the onboard tracking controller, the TG profile was much more uniform, which proved to be very beneficial, especially during braking maneuvers where the operator took much longer to stop the vehicle.
\begin{figure}[!b]
  \vspace*{-6mm}
  \centering
  \includegraphics[trim={0mm 2mm 0mm 4mm},clip,width=1\columnwidth]{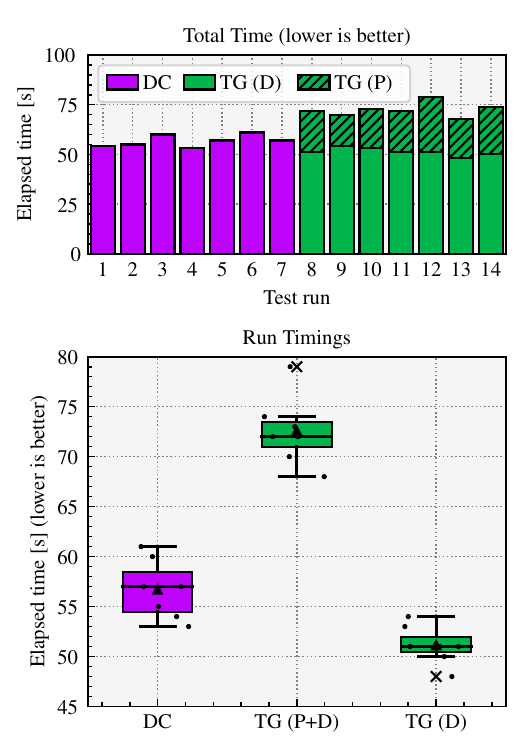}
  \caption{DC vs. TG: Timing results}
  \label{Fig:Timings}
\end{figure}

The time axis reveals the main difference between the two concepts. Using \ac{dc}, the operator starts the teleoperation immediately, while with \ac{tg}, the operator has to spend some time planning the path for the vehicle to follow. Consequently, this slows down the interaction and increases the total teleoperation time. The average total\footnote{The total time was normalized to exclude the time it took for the vehicle to disengage the autonomous operation and transit into teleoperation-ready mode. This cloud-based process took, on average, $35.7$ seconds consistently for all runs.} teleoperation time for DC was $\bar{t}_{DC}=56.7$ s, while TG (P+D) took $\bar{t}_{TG}=72.6$~s which represents an increase of ${\sim}28\%$. The planning time for TG (P), $\bar{t}_{TG^P}$ took on average $21.4$ s. This means that the average time spent on the actual remote driving TG (D) was $\bar{t}_{TG^D}=51.2$ s, which is  ${\sim}10\%$ better (shorter) than the DC despite taking the longer paths (see Fig. \ref{Fig:results_e}). These are very interesting results as they show the operator's ability to plan the path in a reasonable time and that the teleoperation itself could be carried out possibly faster than with direct control, potentially mitigating the effect of \emph{lost} time during trajectory planning. All of that while leaving room for further improvements and optimization of the TG concept. The results of each run are summarized in Fig. \ref{Fig:Timings}.

\subsection{Future Work}
It is important to remember that the TG concept was not optimized for speed but rather for safety. The main idea of relieving the operator of the vehicle stabilization task aims to reduce the cognitive load and increase situational awareness. A user study with a larger group of participants is necessary to assess these aspects in the context of TG. The live demonstration event offered a unique and valuable opportunity to explore the concept ideas in a realistic environment. However, the event was not designed to substitute a user study, and as such, the results should be interpreted as a proof of concept rather than a final evaluation and used as a basis for further research. The identified room for improvement might include path-planning assistance to speed up the planning process, possibly using additional data sources similar to remote assistance concepts such as collaborative planning (see \cite{Majstorovic2023-xf}).

Developed safety mechanisms proved effective but were not tested in a wide range of real-world scenarios. The next step would be to test the system with dynamic traffic participants and more uncertain scene conditions to assess their robustness and reliability. The introduction of a third safety mechanism in the form of a collision avoidance module would be a logical next step to close the loop. Finally, the experimental validation confirmed the importance of the HMI design. As a primary interface between the operator and the vehicle, the HMI is crucial in enabling the operator to perform the teleoperation task. The developed HMI concepts demonstrated the essential functionalities and can be used as indicators of the design decisions for future development cycles. Additionally, the input and output device selection should be carefully evaluated to ensure the best possible user performance and experience.

\label{sec:discussion}

\newpage
\section{Conclusion}
\label{sec:conclusion}
This paper presented a novel technical implementation of the trajectory guidance teleoperation concept. The concept was developed as part of the UNICARagil research project and was experimentally validated using the autoTAXI ride-hailing vehicle prototype. The results indicate that the trajectory guidance concept can be a viable direct control alternative to remotely operate the vehicle in a safe and comfortable manner. Requiring a planning phase before the teleoperation starts, the concept comes with an additional time cost. However, the experimental results indicate that the operator can plan the path in a reasonable time and that remote driving, in specific scenarios, can be carried out possibly faster than with direct control. This opens up the possibility of further concept optimizations.  Assessment of the operator's workload was beyond the scope of this work and should be addressed in future research. In addition to the remote driving task, the technical implementation also included the development and validation of the corresponding safety mechanisms. They proved effective and reliable but also showed the need for further verification in more complex traffic environments.

\addtolength{\textheight}{-12cm}   %

\section*{ACKNOWLEDGMENT}

Domagoj Majstorovic and Simon Hoffmann, as the first authors, were the main developers of the presented work. Frank Diermeyer made essential contributions to the conception of the research project and revised the paper critically for important intellectual content. He gave final approval for the version to be published and agreed to all aspects of the work. As a guarantor, he accepts responsibility for the overall integrity of the paper. The authors want to thank David Brecht, Nils Gehrke, Philipp Hafemann, Andreas Schimpe, and Xulin Song for their support during the experiment phase. The research was financially supported by the Federal Ministry of Education and Research of Germany (BMBF) under grants No. 16EMO0288 (UNICAR.agil) and No. 01IS22088 (AUTOtech.agil).

\bibliographystyle{IEEEtran}
\bibliography{IEEEabrv,references}

\end{document}

%% file: content/acronyms.tex
\begin{acronym}
    \acro{ad}[AD]{Automated Driving}
    \acro{ar}[AR]{Augmented Reality}
    \acro{av}[AV]{Autonomous Vehicle}
    \acro{tc}[TC]{Teleoperation Concept}
    \acro{sa}[SA]{Situational Awareness}
    \acro{dc}[DC]{Direct Control}
    \acro{sc}[SC]{Shared Control}
    \acro{tg}[TG]{Trajectory Guidance}
    \acro{cnn}[CNN]{Convolutional Neural Network}
    \acro{cots}[COTS]{commercial off-the-shelf}
    \acro{cpp}[CPP]{Collaborative Path Planning}
    \acro{dcpp}[DCPP]{Dynamic Collaborative Path Planning}
    \acro{hmi}[HMI]{Human-Machine Interface}
    \acro{ipp}[IPP]{Interactive Path Planning}
    \acro{mrm}[MRM]{Minimal Risk Maneuver}
    \acro{odd}[ODD]{Operational Design Domain}
    \acro{qa}[QA]{Quality Assurance}
    \acro{ra}[RA]{Remote Assistance}
    \acro{rd}[RD]{Remote Driving}
    \acro{rm}[RM]{Remote Monitoring}
    \acro{ros}[ROS]{Robot Operating System}
    \acro{rnd}[R\&D]{research and development}
    \acro{sw}[SW]{Software}
    \acro{tod}[ToD]{Teleoperated Driving}
    \acro{wg}[WG]{Waypoint Guidance}
\end{acronym}

%% file: root.bbl
\begin{thebibliography}{10}
\providecommand{\url}[1]{#1}
\csname url@samestyle\endcsname
\providecommand{\newblock}{\relax}
\providecommand{\bibinfo}[2]{#2}
\providecommand{\BIBentrySTDinterwordspacing}{\spaceskip=0pt\relax}
\providecommand{\BIBentryALTinterwordstretchfactor}{4}
\providecommand{\BIBentryALTinterwordspacing}{\spaceskip=\fontdimen2\font plus
\BIBentryALTinterwordstretchfactor\fontdimen3\font minus
  \fontdimen4\font\relax}
\providecommand{\BIBforeignlanguage}[2]{{%
\expandafter\ifx\csname l@#1\endcsname\relax
\typeout{** WARNING: IEEEtran.bst: No hyphenation pattern has been}%
\typeout{** loaded for the language `#1'. Using the pattern for}%
\typeout{** the default language instead.}%
\else
\language=\csname l@#1\endcsname
\fi
#2}}
\providecommand{\BIBdecl}{\relax}
\BIBdecl

\bibitem{Sexton2023-wb}
B.~Sexton, ``{GM's} cruise robotaxis are {NOT} fully self-driving and require
  humans in remote operations center to intervene on 'tricky drives' every four
  to five miles,''
  \url{https://www.dailymail.co.uk/news/article-12721253/GM-Cruise-robotaxi-not-fully-self-driving.html},
  Nov. 2023, accessed: 2024-2-1.

\bibitem{Majstorovic2022-px}
D.~Majstorovi{\'c}, S.~Hoffmann, F.~Pfab, A.~Schimpe, M.-M. Wolf, and
  F.~Diermeyer, ``Survey on teleoperation concepts for automated vehicles,'' in
  \emph{2022 {IEEE} International Conference on Systems, Man, and Cybernetics
  ({SMC})}, Oct. 2022, pp. 1290--1296.

\bibitem{Amador_undated-yn}
O.~Amador, M.~Aramrattana, and A.~Vinel, ``A survey on remote operation of road
  vehicles.''

\bibitem{noauthor_undated-nr}
``\BIBforeignlanguage{de}{Project {UNICARagil}},''
  \url{https://www.unicaragil.de/}, accessed: 2024-1-13.

\bibitem{Michael_Buchholz_undated-ze}
I.~Michael~Buchholz, F.~Gies, A.~Danzer, M.~Henning, C.~Hermann, M.~Herzog,
  M.~Horn, M.~Sch{\"o}n, N.~Rexin, K.~Dietmayer, C.~Fernandez, J.~Janosovits,
  D.~Kamran, C.~Kinzig, M.~Lauer, E.~Molinos, L.~C. Stiller, S.~Ackermann,
  T.~Homolla, N.~Hermann~Winner, G.~Gottschalg, S.~Leinen, M.~Becker,
  J.~Feiler, S.~Hoffmann, I.~F. Diermeyer, B.~Lampe, T.~Beemelmanns,
  R.~Van~Kempen, T.~Woopen, I.~Lutz, N.~Voget, I.~D. Moormann, I.~Jatzkowski,
  T.~Stolte, M.~Maurer, G.~J{\"u}rgen, I.~E.~V. Hin{\"u}ber, and
  N.~Siepenk{\"o}tter, ``Automation of the {UNICARagil} vehicles,''
  \url{https://oparu.uni-ulm.de/xmlui/bitstream/handle/123456789/34086/D5.3_Buchholz_UniUlm.pdf?sequence=1&isAllowed=y},
  accessed: 2024-1-31.

\bibitem{Lampe2019-vi}
B.~Lampe, L.~Eckstein, and T.~Woopen, ``Collective driving : Cloud services for
  automated vehicles in {UNICARagil},'' 2019.

\bibitem{Georg2019-dt}
J.-M. Georg and F.~Diermeyer, ``An adaptable and immersive real time interface
  for resolving system limitations of automated vehicles with teleoperation,''
  in \emph{2019 {IEEE} International Conference on Systems, Man and Cybernetics
  ({SMC})}, Oct. 2019, pp. 2659--2664.

\bibitem{Schimpe2022-up}
A.~Schimpe, D.~Majstorovic, and F.~Diermeyer, ``Steering action-aware adaptive
  cruise control for teleoperated driving,'' in \emph{2022 {IEEE} International
  Conference on Systems, Man, and Cybernetics ({SMC})}, Oct. 2022, pp.
  988--993.

\bibitem{Kay1995-ab}
J.~S. Kay and C.~E. Thorpe, ``Operator interface design issues in a
  {Low-Bandwidth} and {High-Latency} vehicle teleoperation system,'' \emph{SAE
  Trans. J. Mater. Manuf.}, vol. 104, pp. 487--493, 1995.

\bibitem{Gnatzig2012-bu}
S.~Gnatzig, F.~Schuller, and M.~Lienkamp, ``Human-machine interaction as key
  technology for driverless driving - a trajectory-based shared autonomy
  control approach,'' in \emph{2012 {IEEE} {RO-MAN}: The 21st {IEEE}
  International Symposium on Robot and Human Interactive Communication}, Sep.
  2012, pp. 913--918.

\bibitem{Hoffmann2022-cx}
S.~Hoffmann, D.~Majstorovi{\'c}, and F.~Diermeyer, ``Safe corridor: A
  {Trajectory-Based} safety concept for teleoperated road vehicles,'' in
  \emph{2022 International Conference on Connected Vehicle and Expo
  ({ICCVE})}.\hskip 1em plus 0.5em minus 0.4em\relax IEEE, Mar. 2022, pp. 1--6.

\bibitem{Jatzkowski2021-bx}
I.~Jatzkowski, T.~Stolte, R.~Graubohm, M.~Maurer, A.~Kampmann, B.~Alrifaee,
  S.~Kowalewski, M.~Buchholz, and K.~Dietmayer, ``Integration of a vehicle
  operating mode management into {UNICARagil's} automotive service-oriented
  software architecture,'' 2021.

\bibitem{Kampmann2019-eg}
A.~Kampmann, B.~Alrifaee, M.~Kohout, A.~W{\"u}stenberg, T.~Woopen, M.~Nolte,
  L.~Eckstein, and S.~Kowalewski, ``A dynamic {Service-Oriented} software
  architecture for highly automated vehicles,'' in \emph{2019 {IEEE}
  Intelligent Transportation Systems Conference ({ITSC})}.\hskip 1em plus 0.5em
  minus 0.4em\relax IEEE, Oct. 2019, pp. 2101--2108.

\bibitem{Iso2019-ad}
{ISO}, ``\BIBforeignlanguage{en}{Ergonomics of human-system interaction part
  210: Human-centred design for interactive systems},'' Tech. Rep.
  9241-220:2019, 2019.

\bibitem{Kluge2021-iq}
T.~Kluge, ``Cubic splines,'' 2021.

\bibitem{Lynch_undated-cr}
K.~M. Lynch and F.~C. Park, ``{MECHANICS}, {PLANNING}, {AND} {CONTROL},''
  \url{https://hades.mech.northwestern.edu/images/7/7f/MR.pdf}, accessed:
  2024-1-31.

\bibitem{Homolla2022-jq}
T.~Homolla and H.~Winner, ``Encapsulated trajectory tracking control for
  autonomous vehicles,'' \emph{Automotive and Engine Technology}, vol.~7,
  no.~3, pp. 295--306, Dec. 2022.

\bibitem{Tener2022-yv}
F.~Tener and J.~Lanir, ``Driving from a distance: Challenges and guidelines for
  autonomous vehicle teleoperation interfaces,'' in \emph{{CHI} '22: {CHI}
  Conference on Human Factors in Computing Systems}.\hskip 1em plus 0.5em minus
  0.4em\relax unknown, Apr. 2022, pp. 1--13.

\bibitem{Cruise2021-en}
{Cruise}, ``Cruise under the hood 2021: From {Self-Driving} {R\&D} to
  {Self-Driving} reality,'' Nov. 2021.

\bibitem{Zoox2020-ae}
{Zoox}, ``How zoox uses {TeleGuidance} to provide remote assistance to its
  autonomous vehicles,'' Nov. 2020.

\bibitem{Schimpe2022-ig}
A.~Schimpe, J.~Feiler, S.~Hoffmann, D.~Majstorovi{\'c}, and F.~Diermeyer,
  ``Open source software for teleoperated driving,'' in \emph{2022
  International Conference on Connected Vehicle and Expo ({ICCVE})}, Mar. 2022,
  pp. 1--6.

\bibitem{noauthor_undated-xw}
``\BIBforeignlanguage{en}{Foxglove studio: Robotics visualization and
  debugging}.''

\bibitem{Poggenhans2018-vm}
F.~Poggenhans, J.-H. Pauls, J.~Janosovits, S.~Orf, M.~Naumann, F.~Kuhnt, and
  M.~Mayr, ``Lanelet2: A high-definition map framework for the future of
  automated driving,'' in \emph{2018 21st International Conference on
  Intelligent Transportation Systems ({ITSC})}.\hskip 1em plus 0.5em minus
  0.4em\relax IEEE, Nov. 2018.

\bibitem{noauthor_undated-yo}
``\BIBforeignlanguage{en}{{CarMaker}},''
  \url{https://ipg-automotive.com/en/products-solutions/software/carmaker/},
  accessed: 2024-1-31.

\bibitem{noauthor_undated-st}
``\BIBforeignlanguage{de}{Aldenhoven testing center -- home},''
  \url{https://www.aldenhoven-testing-center.de/de/}, accessed: 2024-1-31.

\bibitem{Majstorovic2023-xf}
D.~Majstorovi{\'c} and F.~Diermeyer, ``Dynamic collaborative path planning for
  remote assistance of {Highly-Automated} vehicles,'' in \emph{2023 {IEEE}
  International Automated Vehicle Validation Conference ({IAVVC})}.\hskip 1em
  plus 0.5em minus 0.4em\relax IEEE, Oct. 2023, pp. 1--6.

\end{thebibliography}
